\documentclass{article}
\usepackage{ijcai25}

\usepackage{times}
\usepackage{soul}
\usepackage{url}
\usepackage[hidelinks]{hyperref}
\usepackage[utf8]{inputenc}
\usepackage[small]{caption}
\usepackage{graphicx}
\usepackage{amsmath}
\usepackage{amsthm}
\usepackage{booktabs}
\usepackage{algorithm}
\usepackage{algorithmic}
\usepackage[switch]{lineno}

\usepackage{colortbl}
\usepackage{inconsolata}

\usepackage{graphicx}
\usepackage[table,dvipsnames]{xcolor}
\usepackage{booktabs}
\usepackage{longtable}
\usepackage{listings}

\usepackage{natbib}
\usepackage{pifont}
\usepackage{amssymb}
\usepackage{multirow}
\usepackage{array}
\usepackage{amsmath}

\usepackage{supertabular}

\title{FinSphere, a Real-Time Stock Analysis Agent Powered by Instruction-Tuned LLMs and Domain Tools}

\author{
Shijie Han$^{1,2}$
\and
Jingshu Zhang$^{1,3}$\and
Yiqing Shen$^{1,4}$\and
Kaiyuan Yan$^1$\and
Hongguang Li\thanks{Corresponding author: \href{mailto:harvey2@mail.ustc.edu.cn}{harvey2@mail.ustc.edu.cn}\\\textit{Accepted by FinLLM @ International Joint Conference on Artificial Intelligence}, Guangzhou, China, 2025. Copyright 2025 by the author(s).}$^1$\\
\affiliations
$^1$JF SmartInvest Holdings Ltd.\\
$^2$Columbia University\\
$^3$Shanghai University of Finance and Economics\\
$^4$Johns Hopkins University\\
}
\begin{document}

\maketitle

\begin{abstract}
Current financial large language models (FinLLMs) struggle with two critical limitations: the absence of objective evaluation metrics to assess the quality of stock analysis reports and a lack of depth in stock analysis, which impedes their ability to generate professional-grade insights. To address these challenges, this paper introduces FinSphere, a stock analysis agent, along with three major contributions: (1) AnalyScore, a systematic evaluation framework for assessing stock analysis quality, (2) Stocksis, a dataset curated by industry experts to enhance LLMs' stock analysis capabilities, and (3) FinSphere, an AI agent that can generate high-quality stock analysis reports in response to user queries. Experiments demonstrate that FinSphere achieves superior performance compared to both general and domain-specific LLMs, as well as existing agent-based systems, even when they are enhanced with real-time data access and few-shot guidance. The integrated framework, which combines real-time data feeds, quantitative tools, and an instruction-tuned LLM, yields substantial improvements in both analytical quality and practical applicability for real-world stock analysis.
\end{abstract}
\section{Introduction}
Large Language Models (LLMs) have demonstrated remarkable capabilities in comprehending and processing natural language, extending their influence across various domains, including finance \citep{li2023large}. By leveraging their language comprehension capabilities, these models have exhibited exceptional performance in various financial applications, including sentiment analysis \citep{liu2024large, zhang2023instruct} and information extraction from unstructured financial texts \citep{li2024extracting, huang2023finbert}. The advent of finance-specific LLMs such as FinBERT \citep{yang2020finbert, liu2021finbert}, BloombergGPT \citep{wu2023bloomberggpt}, and PIXIU \citep{xie2023pixiu} has further enhanced the capacity to process financial data effectively. These advancements have laid the foundation for developing more sophisticated financial analysis tools and shifted how investors interact with market data \citep{krause2023large, nie2024survey}. These AI-powered systems have broadened access to professional financial insights, allowing retail investors to benefit from advanced analysis once reserved for institutions. 

As LLM technology continues to evolve, there is a growing expectation for these models to handle more complex financial tasks, particularly in stock analysis and real-time financial question-answering \citep{yang2023investlm, zhao2024revolutionizing}. This has led to the development of tool-augmented agents that integrate LLMs' natural language understanding with specialized financial tools, significantly enhancing automated financial analysis and interactive question-answering capabilities \citep{ding2024large, zhang2024multimodal}. However, LLM-based financial QA systems still face substantial challenges in effectively interpreting and utilizing the outputs of these tools to generate high-quality analytical responses. Two primary obstacles include the absence of systematic evaluation frameworks to assess their performance in stock analysis, as well as the lack of specialized datasets for fine-tuning LLMs' analytical reasoning capabilities. Moreover, existing research is constrained by LLMs' heavy reliance on historical data, such as GPT-4o's dependence on its pre-trained knowledge for generating responses \citep{ni2024harnessing, bhat2024stock}. This limitation in accessing and processing real-time financial data and domain-specific information restricts their ability to fully capture the dynamic and evolving nature of financial markets, posing a critical challenge for real-time financial question-answering systems.

% To address these limitations, we present three key contributions:  
We address these limitations through three contributions:
\begin{itemize}  
    \item \textbf{AnalyScore}: A comprehensive evaluation framework designed to systematically assess the accuracy, relevance, and analytical depth of LLM-driven stock question-answering.  
    \item \textbf{Stocksis}: A specialized dataset curated by industry experts to enhance LLMs' financial question-answering and stock analysis capabilities. 
    \item \textbf{FinSphere}: A real-time stock question-answering agent capable of generating high-quality stock analysis reports in response to user queries.  
\end{itemize}  
\begin{table*}[ht]
\small
\centering
\begin{tabular}{|p{0.525\textwidth}|p{0.415\textwidth}|}
\hline
\textbf{Prompt with Background Information (input)} & \textbf{Expert-Edited Analysis (label)} \\
\hline
\textbf{System Instruction.}
You are a proficient investment assistant, adept at stock analysis tasks. You demonstrate a robust aptitude for comprehending user-provided \textit{[Question]} and meticulously organizing them into accurate and standardized analysis reports. You utilize the \textit{[Background Information]} provided by the corresponding quantitative tools to respond to users in a systematic and standardized manner.

\vspace{0.1cm}
\textbf{Background Information (from quantitative tools).}

\begin{itemize}
    \item Capital Flow Analysis: TF Securities is currently under moderate control by market leaders, indicating a positive institutional outlook. There is a strong willingness for margin financing inflow. Institutional quarterly shareholding has increased, suggesting institutions are optimistic about the company's development, warranting attention to medium-term opportunities.

    \item \textit{[Other specialized analysis from quantitative tools]}
\end{itemize}

\textbf{Question.}
Please analyze TF Securities.
& 
From a capital flow perspective, the 5-day cumulative DDX of -14.865 indicates significant selling pressure from major market participants and strong bearish sentiment. Despite moderate institutional control and determined margin financing inflows showing relatively active main capital, short-term capital flows warrant caution.

\textit{[Key Summaries of supporting conclusions drawn from other Specialized Analysis]}

\textbf{In conclusion, TF Securities shows recent unusual market activity, largely driven by heightened trading within the securities sector. Technical analysis reveals short-term opportunities with a bullish engulfing pattern, though weak fundamentals warrant caution for medium to long-term positions. Investors should monitor market trends and company developments, as the stock benefits from sector momentum despite underlying fundamental concerns.} \\
\hline
\end{tabular}
\caption{An abbreviated example of Stocksis. The complete content is detailed in Table \ref{tab: tuning_dataset_full}. Guide LLM to provide a comprehensive analysis based on specialized analyses returned from quantitative tools. The average cost of comprehensive analysis written by experts is \$10 per query, with detailed cost breakdowns available in Appendix \ref{sec: cost}.}
\label{tab: tuning_dataset_ex}
\end{table*}

Our experiments demonstrate that FinSphere, by integrating real-time financial databases, specialized quantitative tools, and an instruction-tuned LLM optimized for financial question-answering, significantly outperforms both general-purpose and domain-specific LLMs, as well as existing agent-based systems. This superior performance holds even when baseline models are augmented with real-time background information and few-shot prompting, validating the effectiveness of our integrated approach to real-time financial question-answering and stock market analysis.

\section{AnalyScore and Stocksis}
%Stock market analysis is becoming increasingly complex, necessitating the integration of diverse data sources and sophisticated analytical approaches. While LLMs demonstrate potential in transforming financial analysis, there are two critical gaps in the current landscape: the absence of standardized evaluation frameworks for assessing the quality of AI-generated stock analyses and the lack of high-quality training data for developing LLMs' stock analysis capabilities. This section introduces two significant contributions to address these gaps: AnalyScore, a systematic evaluation framework for assessing stock analysis, and Stocksis, a comprehensive dataset specifically designed to enhance LLMs' stock analysis capabilities. 
Stock market analysis is becoming increasingly complex, necessitating the integration of diverse data sources and sophisticated analytical approaches. While LLMs show promise in financial analysis, two key gaps remain: the lack of standardized evaluation frameworks for AI-generated stock analyses and the scarcity of high-quality training data. To address these, we introduce \textbf{AnalyScore}, a systematic evaluation framework, and \textbf{Stocksis}, a comprehensive dataset designed to enhance LLMs’ stock analysis capabilities.
\subsection{AnalyScore: A Comprehensive Evaluation Framework for Stock Analysis Reports}\label{sec: analyscore}

To address the lack of systematic evaluation standards in stock analysis, we introduce \textbf{AnalyScore}, a domain-specific framework co-developed with industry experts. It combines established financial assessment criteria with insights from LLM-generated content evaluation.

AnalyScore employs a structured scoring system across four dimensions, with a total score of 100 points:

\begin{itemize}
    \item \textbf{Conclusion} (20 pts): Clarity, relevance, and personalization of investment recommendations.
    \item \textbf{Content} (45 pts): Depth, coherence, and professionalism of analytical reasoning.
    \item \textbf{Expression} (15 pts): Organization, fluency, and linguistic clarity.
    \item \textbf{Data} (20 pts): Breadth, accuracy, and effective use of supporting quantitative data.
\end{itemize}

This multi-dimensional evaluation ensures comprehensive coverage of both qualitative judgment and quantitative rigor. At present, AnalyScore is used exclusively by human experts to assess model-generated analyses. In future work, we aim to integrate AnalyScore into automatic evaluation pipelines by prompting LLMs to emulate expert scoring behaviors.

\subsection{Stocksis: A High-Quality Dataset for Enhancing LLMs' Stock Analysis}\label{sec: stocksis}
To evaluate LLMs’ capabilities in stock analysis, we first assess GPT-4o’s responses using AnalyScore (Section~\ref{sec: contrast}), providing it with extensive background knowledge including market data and quantitative indicators. Despite these supports, its outputs often exhibit reasoning inconsistencies, shallow financial insights, and occasional misinterpretations of market trends—revealing the challenges LLMs face in synthesizing complex financial information into coherent, actionable analysis. To address these limitations, we collaborate with industry experts to iteratively refine GPT-4o’s outputs, correcting errors and enriching them with deeper reasoning. This expert-guided process results in Stocksis, a high-quality dataset that bridges automated generation and professional-grade analysis, offering structured supervision to improve LLM performance in real-world financial contexts.

Stocksis comprises 5,000 meticulously curated training pairs, with part of them\footnote{Open-sourced Stocksis has been anonymized and is available at \url{https://anonymous.4open.science/r/Stocksis-BD25/}} available in the open-source release for research and development purposes. An abbreviated example is shown in Table \ref{tab: tuning_dataset_ex}, and the complete content of the same sample is detailed in Table \ref{tab: tuning_dataset_full}. Each training sample consists of two key components:  
\begin{itemize}
\item Prompt with Background Information (input): A complete analytical prompt that includes aggregated outputs from multiple quantitative analysis tools (averaging six tools per sample) as background information. The background information covers volume-price analysis, technical indicators, and other market metrics. Each prompt is rigorously crafted to guide the model in performing analytical tasks while leveraging the provided background information. The average length is 4,000 words.
\item Expert-Edited Analysis (label): In-depth analytical reports responding to the prompt's requirements while effectively utilizing the background information, averaging 3,000 words per report. Due to the particularity of the stock analysis task, there is no standard answer to this task. Therefore, our industry experts provide a high-quality reference analysis result for this task by evaluating the overall market, providing detailed reasons, and demonstrating how to effectively interpret various quantitative indicators.
\end{itemize}

\paragraph{Dataset Collection and Quality Assurance.}
Stocksis is constructed through a structured pipeline grounded in our company’s expertise in retail-oriented stock analysis. Data collection involves two main stages:
\begin{enumerate}
\item \textbf{Prompt and Background Generation}: Expert analysts select suitable quantitative tools for specific stock queries and craft prompts enriched with structured analytical outputs.
\item \textbf{Analysis Refinement}: GPT-4o generates initial responses, which are then iteratively reviewed and improved by a panel of 10 senior analysts to ensure accuracy, coherence, and domain relevance. This process spans over three months to ensure high-quality outputs.
\end{enumerate}

Stocksis addresses a key gap in financial NLP: the lack of datasets combining structured prompts with expert-refined reasoning. Unlike existing resources focused on price or sentiment data, it enables the fine-tuning of general-purpose LLMs to enhance their capability in structured, high-quality financial analysis for real-world applications.

\section{FinSphere Agent}\label{sec: method}
This section details the architecture and operational mechanisms of FinSphere Agent, our advanced stock analysis agent.
\begin{figure}[ht]
\centering
\includegraphics[width=\linewidth]{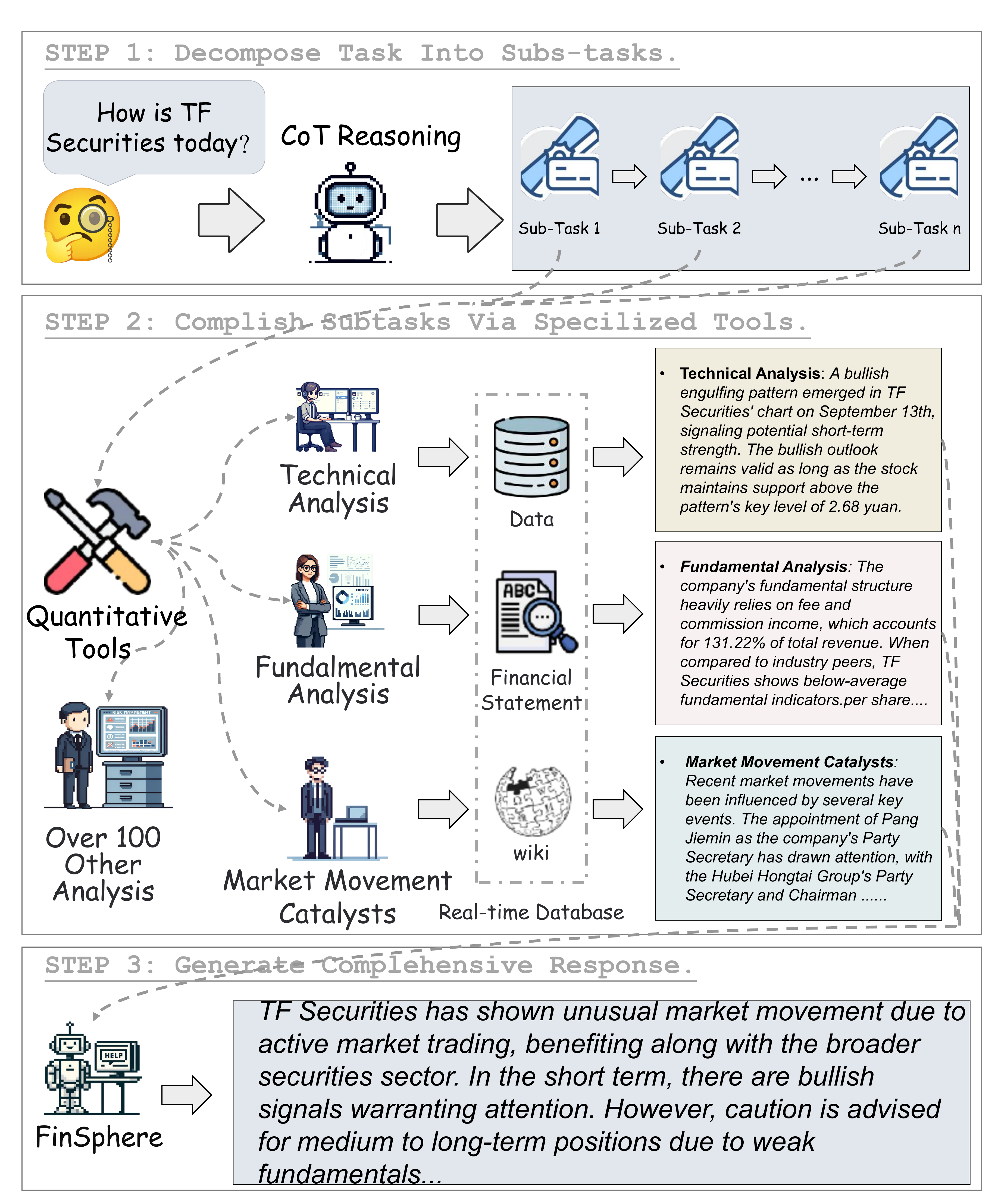}
\caption{This diagram illustrates the overall workflow of the FinSphere Agent, detailing how different components interact to facilitate real-time stock analysis.}
\label{fig: finsphere}
\end{figure}

\begin{table*}
\tiny
\resizebox{\textwidth}{!}{%
\begin{tabular}{l|c|c|c|c|c}
\hline
 & Conclusion & Content & Expression & Data & Total\\
 & (Score: 20) & (Score: 45) & (Score: 15) & (Score: 20) & Score\\
\hline
GPT-4o  & 9.85 & 26.12 & 12.44 & 18.20 & 66.61\\
Deepseek-v3  & 9.52 & 25.30 & 12.75 & 16.85 & 64.42 \\
GPT3.5  & 7.95 & 21.05 & 10.15 & 14.30 & 53.45\\
Qwen2-72B  & 8.15 & 22.55 & 10.55 & 14.95 & 56.20\\
InvestLM & 8.40 & 23.10 & 11.25 & 15.75 & 58.50\\
FinGPT  & 6.80 & 18.55 & 8.95 & 10.75 & 40.05\\
\hline
FinRobot  & 9.10 & 24.05 & 11.55 & 16.35 & 61.05\\
FinMem  & 9.90 & 25.95 & 12.85 & 18.85 & 67.55\\
\hline
\rowcolor{blue!25}
FinSphere  & \textbf{9.95} & \textbf{27.16} & \textbf{14.87} & \textbf{18.90} & \textbf{70.88}\\
\hline
\end{tabular}%
}
\caption{Human experts use AnalyScore to evaluate 100 responses generated by 8 models. The scores shown are averages across 100 evaluations. The average cost for expert evaluation is \$10 per response, with detailed cost breakdowns available in Appendix \ref{sec: cost}.}
\label{tab: main}
\end{table*}
\subsection{Powerful Quantitative Tools based on Real-Time Databases}\label{sec: tools}
A core strength of FinSphere lies in its seamless integration with our company's mature suite of quantitative analysis tools, which have been extensively deployed and validated in production environments. These tools access our comprehensive real-time financial database, which maintains extensive coverage of market stocks, including both structured data (price movements, trading volumes, financial metrics) and unstructured data (corporate announcements, analyst reports, market news).

When FinSphere identifies the necessity for specific quantitative analysis, it triggers the corresponding tool from our production suite. These tools then automatically query our real-time database to extract the most recent relevant data, perform sophisticated calculations, and generate specialized analyses such as technical indicators, fundamental valuations, or market sentiment assessments. Each tool is designed to provide contextual information specifically tailored to user queries, leveraging our continuously updated database to ensure that all analyses accurately reflect current market conditions. This architecture ensures that FinSphere's responses are always grounded in the most recent market data while benefiting from our proven quantitative methodologies.

\subsection{Instruction Tuning}
We adopt full-parameter \textbf{instruction tuning} to adapt Qwen2-72B into a domain-specialized financial analysis agent. Using our expert-curated Stocksis dataset—comprising 5,000 structured training pairs of quantitative tool outputs and corresponding expert-written analyses (Section~\ref{sec: stocksis})—we guide the model to learn not just domain knowledge, but also the task format, reasoning flow, and response style required for professional stock reporting. Unlike parameter-efficient methods, full instruction tuning enables the model to internalize complex analytical patterns and generate coherent, multi-dimensional reports.

Training is performed on an NVIDIA 16$\times$A100 GPU cluster with 32K token context, using a language modeling objective with a learning rate of 1e-5, batch size of 16, and 2 training epochs. We apply gradient clipping and mixed-precision training to ensure numerical stability and efficiency. Through this process, FinSphere acquires strong capabilities in synthesizing diverse financial signals, interpreting market data accurately, and responding with structured, human-level stock analyses tailored to real-world scenarios.

\subsection{Overall Workflow}
FinSphere follows a structured, multi-stage workflow to generate real-time financial analyses. Upon receiving a user query, it first applies chain-of-thought (CoT) reasoning to decompose the task into interpretable subtasks and determine which domain-specific quantitative tools are needed. These tools then independently access a real-time financial database to retrieve the latest market data, including technical indicators, capital flows, and fundamental metrics, each generating specialized outputs tailored to their analytical focus.

In the final stage, FinSphere’s instruction-tuned LLM—trained on the Stocksis dataset—serves as an expert analyst. It synthesizes the multi-source outputs into a coherent, structured report aligned with professional financial standards. This integrated architecture enables FinSphere to combine the precision of automated quantitative analysis with the contextual depth of expert reasoning, ensuring responses are both analytically sound and up-to-date.
\section{Evaluation}
Given FinSphere's integration with real-time financial databases and proprietary quantitative tools, it possesses analytical capabilities that extend beyond those of general-purpose LLMs. Performance comparisons between FinSphere and general LLMs present inherent challenges, primarily due to the latter's inability to access real-time financial data and domain-specific information. For example, GPT-4o typically acknowledges its limitations with responses like \textit{"As an AI language model with knowledge cut-off in June 2024, I don't have access to real-time stock information."} To demonstrate FinSphere's enhanced capabilities while ensuring a fair comparison, we have implemented a comprehensive experimental design.
\begin{table*}
\centering
\resizebox{\textwidth}{!}{%
\begin{tabular}{lrrrrrrr}
\toprule
{} &  Group 1 \& 2 &  Group 1 \& 3 &  Group 1 \& 4 &  Group 2 \& 3 &  Group 2 \& 4 &  Group 3 \& 4 &  Average \\
\midrule
Conclusion &        79.36 &        73.90 &        79.51 &        81.59 &        85.30 &        89.63 &    81.55 \\
Content    &        93.77 &        71.45 &        94.25 &        77.61 &        73.49 &        74.99 &    80.93 \\
Expression &        88.30 &        91.65 &        90.81 &        83.12 &        77.30 &        82.86 &    85.67 \\
Data       &        84.97 &        85.03 &        75.31 &        80.80 &        79.16 &        84.81 &    81.68 \\
Total      &        78.90 &        87.70 &        74.55 &        77.28 &        81.40 &        73.16 &    78.83 \\
\bottomrule
\end{tabular}%
}
\caption{Average Kendall’s Tau across 100 queries for different annotator groups rating. Reported as \%}
\label{tab: kendall_tau}
\end{table*}
\paragraph{Baseline.}
We evaluate three categories of models: (1) single LLMs, including proprietary (GPT-4o, GPT-3.5), open-source (Qwen2-72B, Deepseek-v3 \citep{liu2024deepseek}), and domain-specific models (InvestLM \citep{yang2023investlm}, FinGPT \citep{yang2023fingpt}); (2) agent-based systems, including FinMem \citep{yu2024finmem} and FinRobot \citep{yang2024finrobot}; and (3) our proposed FinSphere. All LLMs use chain-of-thought prompting with few-shot examples and relevant background information (Appendix~\ref{sec: It_prompts}), while agents receive simplified prompts similar to Stocksis's inputs. FinSphere is evaluated directly via user queries, utilizing its integrated tools (Appendix~\ref{sec: test_query}). For all models, we set a maximum output length of 8K tokens and a temperature of 0.5.

\subsection{Performance Analysis} \label{sec: contrast}
The expert evaluation results presented in Table \ref{tab: main} demonstrate FinSphere’s superior performance across all assessment dimensions, achieving an overall score of 70.88 out of 100. This surpasses both traditional LLM-based approaches and other agent-based systems, with FinMem and GPT-4o following at 67.55 and 66.61, respectively. The evaluation reveals a clear performance hierarchy: agent-based systems generally outperform standalone language models (except GPT-4o), while general-purpose LLMs show moderate performance and domain-specific LLMs such as FinGPT (40.05) demonstrate relatively limited capabilities. These results validate the effectiveness of FinSphere’s integrated approach, which combines real-time data access, quantitative tools, and a Stocksis-tuned LLM, enabling more precise and insightful stock analysis. More analysis is shown in Appendix \ref{sec:more_ana}.

\subsection{Evaluation Consistency}
We assess the reliability of human evaluation using Kendall’s Tau rank correlation to measure inter-group agreement. Forty industry experts are divided into four groups of ten, each producing consensus scores for all model responses. Pairwise Kendall’s Tau is computed across groups based on rankings of 100 queries (Table~\ref{tab: kendall_tau}).

Results show strong agreement, with most correlations exceeding 80\% and values ranging from 71.45 to 94.25. While Content shows the most variation, Expression and Data maintain stable consistency. These findings indicate that, despite inherent subjectivity, our evaluation process yields robust and reliable assessments of model performance.

\subsection{Ablation Study}
\begin{figure}[ht]
\centering
\includegraphics[width=\linewidth]{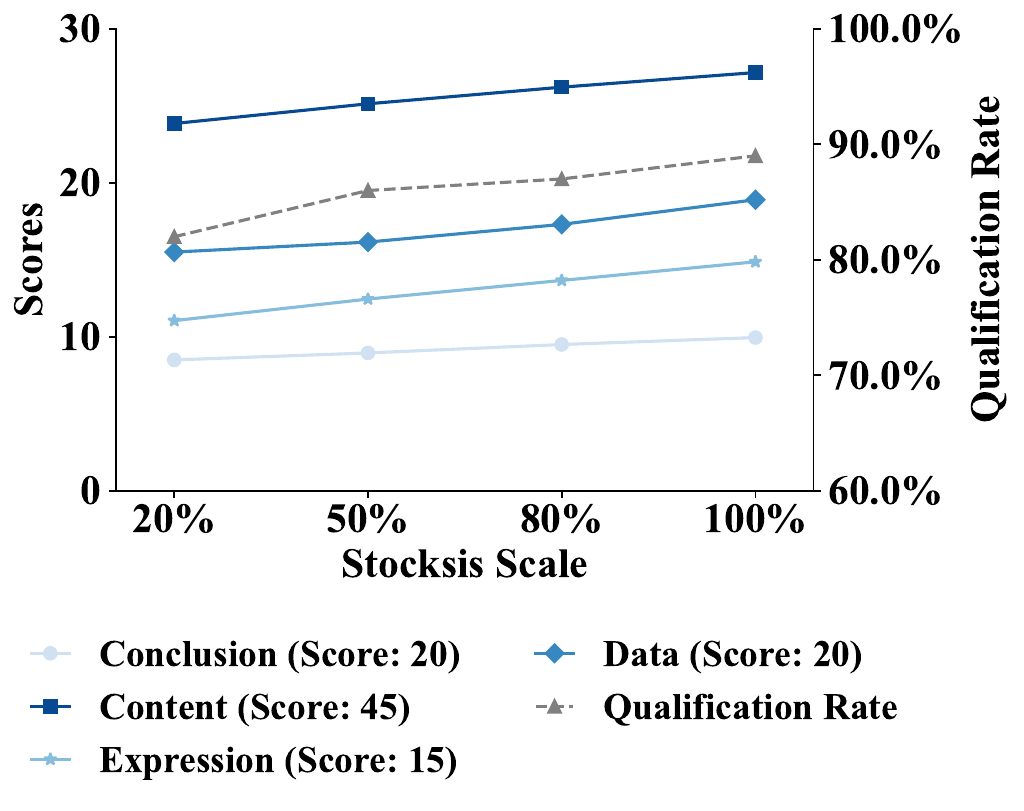}
\caption{Changes in scores of each sub-item as the number of Stocksis used for fine-tuning changes.}
\label{fig: abl}
\end{figure}
To investigate the impact of training data scale on FinSphere's performance, we conduct an ablation study using different proportions of the Stocksis dataset. We fine-tune Qwen2-72B using 20\%, 50\%, 80\%, and 100\% of the 5,000 data pairs while maintaining FinSphere's framework. The detailed evaluation results are provided in Table \ref{fig: abl}, which demonstrates a clear positive correlation between training data scale and model performance, with overall scores increasing from 58.90 (20\%) to 70.88 (100\%). The performance improvement shows a non-linear pattern, with larger incremental gains observed at higher data volumes. These findings underscore the importance of training data in achieving optimal performance, while also demonstrating the robustness of our framework, as FinSphere maintains satisfactory performances even with reduced training data scale.

\section{Related Works}
LLMs have shown strong potential in financial tasks such as stock prediction, market analysis, and portfolio management \citep{zhao2024revolutionizing,li2023large,ni2024harnessing,bhat2024stock,wu2024portfolio,kim2024financial}. Domain-specific models like InvestLM \citep{yang2023investlm} and GPT-InvestAR \citep{gupta2023gpt} further highlight the benefits of financial instruction tuning. LLMs have also been applied to financial anomaly detection \citep{park2024enhancing} and financial statement understanding \citep{kim2024financial}.

\paragraph{Financial Datasets and Evaluation}
Existing datasets such as FinQA \citep{chen2021finqa}, TAT-QA \citep{zhu2021tat}, and FLARE \citep{xie2023pixiu} focus on question answering or numerical reasoning, but lack support for comprehensive, real-time stock analysis. Others, including FinTextQA \citep{chen2024fintextqa}, CFBenchmark \citep{lei2023cfbenchmark}, and FinanceBench \citep{islam2023financebench}, provide broader coverage but miss expert-annotated stock reports. Evaluation of financial text has typically relied on generic metrics (e.g., BLEU \citep{papineni2002bleu}, ROUGE \citep{rouge2004package}), while recent efforts like FinEval \citep{zhang2023fineval} offer domain-adapted metrics but lack systematic expert-grounded scoring.

\paragraph{Instruction Tuning and Tool-Augmented Agents}
Recent advances integrate financial LLMs with domain-specific tools, such as FinGPT \citep{yang2023fingpt}, XBRL-Agent \citep{han2024xbrl}, and FinOps \citep{li2023finops}, improving task relevance and interactivity. However, most rely on static historical data and lack real-time adaptability. In contrast, our proposed FinSphere (Section~\ref{sec: method}) combines an instruction-tuned LLM with real-time databases and quantitative tools for dynamic and actionable stock analysis.
\section{Conclusion}
This paper introduces FinSphere, an innovative stock analysis agent that addresses critical gaps in the capabilities of LLMs for stock analysis. By integrating real-time financial databases, quantitative tools, and an instruction-tuned LLM, FinSphere demonstrates superior performance in generating comprehensive stock analyses. The development and release of Stocksis, a high-quality dataset for enhancing LLMs' stock analysis capabilities, and AnalyScore, a systematic evaluation framework, provide valuable resources for advancing research in AI-powered financial analysis. Our experimental results indicate that FinSphere consistently outperforms general-purpose, domain-specific LLMs and Agent systems across multiple evaluation dimensions, highlighting the effectiveness of our integrated approach.

\appendix

\clearpage
\section{AnalyScore Details}
AnalyScore, introduced in Section \ref{sec: analyscore}, is fully presented in this section, hoping to promote related research in academia and industry. Please refer to Table \ref{tab: analyscore_d} for the specific composition of AnalyScore.

\section{Dimensional Analysis and Visualization}\label{sec:more_ana} 
\begin{figure}[ht]
\centering
\includegraphics[width=\linewidth]{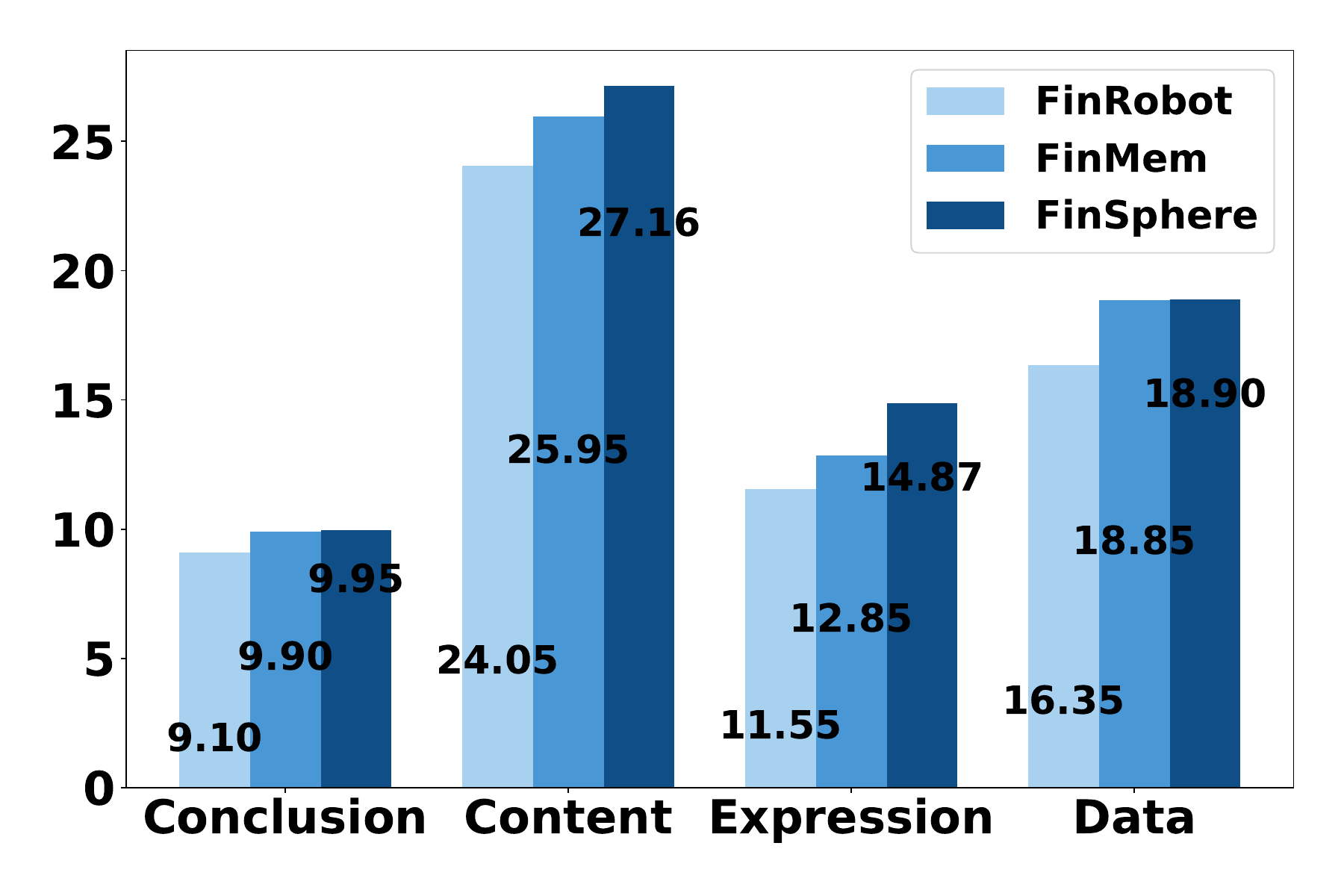}
\caption{Comparison of FinSphere and Agent-based systems in various dimensions.}
\label{fig: res}
\end{figure}

To further investigate the comparative strengths of FinSphere, we conduct a detailed analysis of its performance relative to two other leading agent-based systems, FinRobot and FinMem. The comparative visualization in Figure \ref{fig: res} highlights performance differences across four critical dimensions: Conclusion, Content, Expression, and Data capabilities.

From the visualization, FinSphere exhibits the highest scores across all dimensions. In the Conclusion category, the three models perform relatively closely, with FinSphere slightly leading at 9.95, followed by FinMem at 9.90 and FinRobot at 9.10, demonstrating their robust ability to derive investment insights. However, in the Content dimension, FinSphere shows a clear advantage, scoring 27.16, significantly surpassing FinMem (25.95) and FinRobot (24.05), reflecting its greater analytical depth and content richness.

The most pronounced gap is observed in the Expression dimension, where FinSphere achieves 14.87, noticeably higher than FinMem (12.85) and FinRobot (11.55). This highlights FinSphere’s superior logical articulation in financial reporting. In terms of Data utilization, FinSphere (18.90) and FinMem (18.85) exhibit comparable performance, both substantially outperforming FinRobot (16.35), reinforcing their accurate understanding and grasp of data in financial analysis

On the other hand, one notable limitation of general-purpose LLMs is their heavy dependence on extensive in-context examples to generate accurate financial analyses. This results in a substantial increase in input token consumption, leading to higher operational costs for models such as GPT-4o, and restricting the usability of small context-window LLMs. In contrast, FinSphere’s instruction-tuned architecture eliminates the need for verbose prompts, allowing it to generate high-quality outputs with significantly fewer input tokens.

The combined findings underscore FinSphere’s state-of-the-art capabilities in stock analysis, driven by its robust data processing and structured analytical reasoning. These results further validate its advantage over both standalone LLMs and agent-based financial analysis systems. Additionally, FinSphere has been released to the public with free access in December 2024, with detailed release information provided in Appendix \ref{sec: product}.

\section{Testing Prompt}\label{sec: It_prompts}
As a professional investment advisor, you will analyze stock based on the following:\\

[BACKGROUND INFORMATION - INSERT QUANTITATIVE TOOL RETURNS HERE]\\

[RESPONSE STANDARDS - SEE BELOW]\\

[WRITING GUIDELINES - SEE BELOW]\\

[FEW-SHOTS]\\

[SPECIFIC QUERY]\\

\subsection{Response Standards}

\underline{Step 1: Movement Summary}
\begin{itemize}
\item Clearly state whether stock shows unusual movement
\item Summarize the main reason from background information in under 20 words
\end{itemize}

\noindent\underline{Step 2: Conclusions}
\begin{enumerate}
\item Short-term Conclusion (based on technical analysis):
    \begin{itemize}
    \item For bearish signals: suggest caution, observation, risk awareness, avoidance, position control
    \item For bullish signals: suggest appropriate attention, tracking, validation
    \item Note: State conclusion only, no explanation needed
    \end{itemize}

\item Medium/Long-term Conclusion (based on fundamental analysis):
    \begin{itemize}
    \item For bearish signals: suggest caution, observation, risk awareness, avoidance, position control
    \item For bullish signals: suggest appropriate attention, tracking, validation
    \item Note: State conclusion only, no explanation needed
    \end{itemize}

\item Note: Technical analysis showing bullish keywords requires only direct bullish conclusions. Bullish keywords and bearish keywords are shown in the writing guidelines.
\end{enumerate}

\noindent\underline{Step 3: Detailed Analysis}
\\

Provide an overview of the specific content in [background information] (such as volume and price situation, technical aspect, capital aspect, fundamental aspect, and news aspect) as a support for generating conclusions, and cannot change the financial data performance in the background information.
The analysis mode that can be used for reference is as follows:
\begin{itemize}
\item Volume and Price Analysis: Current price, price movement, industry comparison, index comparison, turnover rate, trading volume/value, market comparison. Include specific data and brief commentary.

\item Technical Analysis: Technical patterns, indicators with specific values. Note specialized indicators if present (e.g., "AI Top/Bottom", "Bull Institution signals").

\item Capital Flow Analysis: 5-day cumulative DDX and capital flow data analysis. Include specific DDX values.

\item Fundamental Analysis: Financial and fundamental data evaluation. Include specific values.

\item News Analysis: Latest news and movement causes, including specific numerical data.
\end{itemize}

\noindent\underline{Step 4: Final Summary}
\\

Restate short-term and medium/long-term conclusions with reasons.

\subsection{Writing Guidelines}
\begin{itemize}
\item Provide direct conclusions and analysis, maintain concise response
\item Avoid phrases: "according to", "information shows", "recent performance", "current situation", "comprehensive analysis", "buy", "sell", "hold", "clear position", "build position", "reduce position", "increase position"
\item Minimize transition words: "but", "however", "then", "finally"
\item Maintain natural flow between paragraphs
\item Bullish keywords: bull point, golden cross, strong uptrend, strengthening, active holding, bullish combination, active attention
\item Bearish keywords: bear point, death cross, weakening, cautious observation, bearish combination, weak adjustment, weak decline.
\end{itemize}

\begin{table*}[t]
\setlength\tabcolsep{4pt}
\renewcommand{\arraystretch}{1.1}
\footnotesize
\centering
\begin{tabular}{|p{3.5cm}|p{3cm}|p{7.8cm}|c|}
\hline
\textbf{Evaluation Dimensions} & \textbf{Subdimensions} & \textbf{Specific Standards} & \textbf{Scores} \\
\hline
\multicolumn{2}{|l|}{\multirow{4}{*}[-0.5em]{\begin{tabular}[c]{@{}l@{}}Conclusion\\(Total Score: 20)\end{tabular}}} & 
\begin{tabular}[c]{@{}l@{}}Generates personalized conclusions based on investment\\preferences and risk profiles\\
Provides explicit conclusions tailored to user personas\\
Implements differentiated investment strategies\\
Aligns with risk tolerance and investment horizons\end{tabular} & 20 \\
\cline{3-4}
\multicolumn{2}{|l|}{} & \begin{tabular}[c]{@{}l@{}}Non-personalized but comprehensive analysis\\
Covers diverse investment styles (conservative to aggressive)\\
Enables user self-selection of strategies\\
Shows broad analytical framework applicability\end{tabular} & 10 \\
\cline{3-4}
\multicolumn{2}{|l|}{} & \begin{tabular}[c]{@{}l@{}}Lacks personalization elements\\
Single investment style analysis\\
Limited analytical perspective\\
Insufficient consideration of preferences\end{tabular} & 5 \\
\cline{3-4}
\multicolumn{2}{|l|}{} & No conclusive elements present & 0 \\
\hline

\multirow{10}{*}[-1em]{\begin{tabular}[c]{@{}l@{}}Content\\(Total Score: 45) \\\end{tabular}} & 
\multirow{8}{*}[-1em]{\begin{tabular}[c]{@{}l@{}}Analysis\\Dimensions\\ \end{tabular}} & 
\begin{tabular}[c]{@{}l@{}}Leverages interaction history for personalization\\
Includes $\leq$ 2 non-personalized dimensional analyses\\
Shows accuracy and forward-looking insights\end{tabular} & 30 \\
\cline{3-4}
& & \begin{tabular}[c]{@{}l@{}}Uses interaction history for personalization\\
Provides accurate, targeted content\\
Presents actionable recommendations\end{tabular} & 25 \\
\cline{3-4}
& & \begin{tabular}[c]{@{}l@{}}Non-personalized analysis across $\leq$ 5 dimensions\\
Demonstrates analytical accuracy and logical rigor\end{tabular} & 20 \\
\cline{3-4}
& & \begin{tabular}[c]{@{}l@{}}Analysis across 4 core dimensions\\
Includes real-time market analysis\\
Maintains accuracy and timeliness\end{tabular} & 18 \\
\cline{3-4}
& & \begin{tabular}[c]{@{}l@{}}Analysis across 4 core dimensions\\
Maintains analytical accuracy\end{tabular} & 15 \\
\cline{3-4}
& & \begin{tabular}[c]{@{}l@{}}Analysis across 3 core dimensions\\
Demonstrates accuracy\end{tabular} & 10 \\
\cline{3-4}
& & \begin{tabular}[c]{@{}l@{}}Analysis across 2 core dimensions\\
Maintains basic accuracy\end{tabular} & 5 \\
\cline{3-4}
& & \begin{tabular}[c]{@{}l@{}}Single-dimensional analysis\\
Limited but accurate content\end{tabular} & 0 \\
\cline{2-4}
& \multirow{2}{*}[-0.5em]{\begin{tabular}[c]{@{}l@{}}Logical Consistency\\ \\\end{tabular}} & 
Logical consistency across all components & 15 \\ 
\cline{3-4}
& & Exhibits logical inconsistencies & 0 \\
\hline

\multirow{6}{*}[-0.5em]{\begin{tabular}[c]{@{}l@{}}Expression\\(Total Score: 15)\\\end{tabular}} & 
\multirow{2}{*}[-0.5em]{\begin{tabular}[c]{@{}l@{}}Structure\\ \\ \end{tabular}} & 
Systematic organization & 5 \\ 
\cline{3-4}
& & Lacks structure & 0 \\
\cline{2-4}
& \multirow{4}{*}{\begin{tabular}[c]{@{}l@{}}Language\\ \end{tabular}} & 
\begin{tabular}[c]{@{}l@{}}Clear communication with hierarchical emphasis\\
Logically rigorous with insightful perspectives\\
In-depth analytical content\end{tabular} & 10 \\
\cline{3-4}
& & \begin{tabular}[c]{@{}l@{}}Clear content with essential coverage\\
Basic logical flow\\
Superficial analysis\end{tabular} & 8 \\
\cline{3-4}
& & \begin{tabular}[c]{@{}l@{}}Comprehensible but unclear expression\\
Discontinuous logic\\
Surface-level analysis\end{tabular} & 5 \\
\cline{3-4}
& & \begin{tabular}[c]{@{}l@{}}Unclear and difficult to comprehend\\
Logical inconsistencies\\
Lacks substantive content\end{tabular} & 0 \\
\hline

\multicolumn{2}{|l|}{\multirow{5}{*}[-0.5em]{\begin{tabular}[c]{@{}l@{}}Data\\(Total Score: 20)\end{tabular}}} & 
Extensive data (>3 dimensions) & 20 \\
\cline{3-4}
\multicolumn{2}{|l|}{} & Substantial data (3 dimensions) & 15 \\
\cline{3-4}
\multicolumn{2}{|l|}{} & Moderate data (2 dimensions) & 10 \\
\cline{3-4}
\multicolumn{2}{|l|}{} & Limited data (1 dimension) & 5 \\
\cline{3-4}
\multicolumn{2}{|l|}{} & No data utilization & 0 \\
\hline
\end{tabular}
\caption{Detailed components of AnalyScore}
\label{tab: analyscore_d}
\end{table*}

\section{LLMs' Response to users' query}\label{sec: other_response}
Here we show the responses of four LLMs (using different methods) to real user inquiries, as well as the scores and comments of experts after evaluation using AnalyScore.

\subsection{FinSphere}
Omat Advanced Materials shows unusual movement, mainly due to the company's plan to build Mingyue Lake Industrial Park in Shaoguan High-tech Zone. In the short term, Omat's stock price shows strong momentum, warranting active attention. In the medium to long term, given the company's weak fundamentals, caution is advised.

Regarding price and volume, Omat's latest stock price is 25.28 yuan, up 14.60\%. It ranks 9th among semiconductor stocks, with the sector index up 5.91\% and the Shanghai Composite Index up 0.09\%, significantly outperforming both the sector and the broader market. Omat's turnover rate is 54.01\%, ranking 1st in the semiconductor industry, with total trading volume of 172,900 lots and total trading value of 407 million yuan, ranking 63rd in the industry. While Omat shows strong price performance today with high market participation, trading volume is relatively average.

Technical analysis shows Omat's RSI formed a golden cross above 50 on November 5th, followed by a MACD golden cross above the zero axis on November 8th. The short-term RSI has reached 80.54, indicating overbought conditions, suggesting a possible strong adjustment in the near term.

Regarding capital flows, Omat's 5-day cumulative DDX is 6.397, indicating strong recent institutional buying. The stock shows moderate institutional control, continued margin trading inflows, and increased quarterly institutional holdings, suggesting positive capital sentiment.

Financially, Omat's Q3 2024 report shows significant declines in revenue and net profit attributable to shareholders. The company's solvency is weak, with declining current and quick ratios and steadily rising debt-to-asset ratio, showing poor fundamentals compared to industry peers.

On the news front, Omat acknowledged at its performance briefing that due to economic conditions, the company's performance is under pressure, with revenue down 22.63\% and net profit down 77.38\% in the first three quarters. The company's plan to build Mingyue Lake Industrial Park in Shaoguan High-tech Zone to expand upstream high-purity materials and copper alloy new materials business demonstrates confidence in future development.

In conclusion, Omat shows positive short-term technical indicators with RSI and MACD golden crosses warranting attention. For the medium to long term, given weak fundamentals, poor performance, and weak solvency, caution is advised. Meanwhile, the Bull Force Index signals an entry opportunity, suggesting investors watch for potential trading opportunities.

\paragraph{Score and comments from experts using AnalyScore.}
FinSphere's analysis clearly identifies the causes of unusual price movements and investment recommendations with complete data citations, including sector comparisons. The conclusions match the data: based on strong technical indicators but weak fundamentals, it provides short-term focus and mid-term cautious recommendations. The analysis across different dimensions is logical with sufficient data support. Specific scores are as follows:
\begin{itemize}
    \item Conclusion (10/20):
        \begin{itemize}
            \item Clear distinction between short-term and mid-to-long-term recommendations
            \item Data-driven investment advice
            \item Lacks personalized recommendations
        \end{itemize}
    \item Content (35/45):
        \begin{itemize}
            \item Covers 5 core dimensions (volume/price, technical, capital flow, financial, news)
            \item Professional and in-depth analysis across dimensions
            \item Strong logical consistency
            \item Lacks personalization
        \end{itemize}
    \item Expression (15/15):
        \begin{itemize}
            \item Clear structure with hierarchical organization
            \item Accurate use of professional terminology
            \item Deep analysis with insights
        \end{itemize}
    \item Data (20/20):
        \begin{itemize}
            \item Comprehensive data dimensions (>3 dimensions)
            \item Accurate data citations with comparative analysis
        \end{itemize}
    \item Total: 80
\end{itemize}

\subsection{GPT-4o with few-shot}
Omat Advanced Materials shows unusual movement due to its planned Mingyue Lake Industrial Park project in Shaoguan High-tech Zone, which has boosted market expectations. The stock deserves moderate attention in both short and medium terms.

Regarding price and volume, Omat's latest price is 25.28 yuan, up 14.60\%, ranking 9th in the semiconductor industry. The semiconductor sector index stands at 2,740.03, up 5.91\%. The Shanghai Composite Index is at 3,455.28, up 0.09\%. Omat's turnover rate of 54.01\% ranks first in the industry, with total trading volume of 172,900 lots and trading value of 407 million yuan, ranking 63rd. The stock shows strong price gains and high market activity.

Technically, Omat formed an RSI golden cross above 50 on November 5th, followed by a MACD golden cross above the zero axis on November 8th, maintaining strong momentum. The current RSI of 80.54 indicates overbought conditions, suggesting potential near-term adjustment.

Capital flow analysis shows a 5-day cumulative DDX of 6.397, indicating strong institutional buying and moderate institutional control, with decisive margin trading inflows.

Fundamentally, Omat's Q3 2024 report shows revenue of 297 million yuan for the first three quarters, down 22.63\% year-over-year, and net profit of 9.8746 million yuan, down 77.38\%. The company's solvency is weak, with current ratio down 11.42\%, quick ratio down 12.27\%, and steadily rising debt-to-asset ratio.

Recent news indicates the company's plan to build Mingyue Lake Industrial Park in Shaoguan High-tech Zone, expanding upstream operations, demonstrating confidence in future development. The market views positively the company's multi-channel funding approach for steady project implementation.

In conclusion, Omat warrants moderate short-term attention given technical golden crosses and strong momentum. Medium-term moderate attention is advised, as despite expansion plans, fundamental data remains weak, requiring careful monitoring.

\paragraph{Score and comments from experts using AnalyScore.}
GPT-4o with few-shot's analysis begins with vague conclusions: "worth moderate attention in short term, should pay appropriate attention in medium to long term" lacks a clear stance; while data citations are complete, the analysis lacks depth; technical analysis is oversimplified and fails to fully interpret indicator meanings; the conclusion section lacks sufficient reasoning for investment recommendations. Specific scores are as follows:

\begin{itemize}
    \item Conclusion (10/20):
\begin{itemize}
    \item Ambiguous conclusions
    \item Lack of specific investment recommendations
    \item Insufficient reasoning
\end{itemize}

\item Content (25/45):
\begin{itemize}
    \item Relatively complete dimensional coverage
    \item Insufficient depth of analysis
    \item Simplified technical interpretation
\end{itemize}

\item Expression (15/15):
\begin{itemize}
    \item Basic clear structure
    \item Plain expression
    \item Lacks insights
\end{itemize}

\item Data (15/20):
\begin{itemize}
    \item Complete data citations
    \item Insufficient data interpretation
\end{itemize}
\item Total: 65
\end{itemize}

\subsection{QWen2-72B with few-shot}
Fundamentally, Omat Advanced Materials reported revenue of 297 million yuan for the first three quarters of 2024, down 22.63\% year-over-year, with net profit of 9.8746 million yuan, down 77.38\%. Profitability metrics are weak, with ROE at 1.5\%, net profit margin at 3.33\%, and gross margin at 18.87\%. Valuation metrics show P/E at 264.02 and P/B at 4.12, ranking at 46.5\% and 64.56\% in the industry respectively.

Technical analysis shows significant price volatility with 10-day amplitude and range at 48.21\% and 19.54\% respectively. The 5-day, 10-day, and 20-day moving averages are 21.72, 21.87, and 21.87 yuan respectively. While price is above these averages, their bearish alignment suggests potential near-term pressure.

Capital flow analysis shows volatile institutional flows over the past 10 days, with overall net outflows. Margin trading balance stands at 7.0404 million yuan, representing 0.44\% of float, indicating low leverage participation.

As a STAR Market-listed company, Omat has raised funds for projects including high-end sputtering target production facilities. The company's products serve various sectors including semiconductor displays, touch screens, architectural glass, decorative coating, IC packaging, solar photovoltaic, and power batteries.

In conclusion, Omat requires caution in the short term given bearish technical alignment and weak capital flows. Medium to long-term focus should be on performance improvement and industry development opportunities. Investors should closely monitor company performance, industry trends, and market capital flows.

\paragraph{Score and comments from experts using AnalyScore.}
QWen2-72B with few-shot's analysis lacks explanation of the causes behind unusual movements; technical analysis is disconnected from current trends and fails to reflect recent changes; capital flow descriptions are too general without specific data support; news analysis is superficial and fails to demonstrate deep analysis of company strategy; conclusions are overly conservative and don't match some positive signals. Specific scores are as follows:

\begin{itemize}
    \item Conclusion (5/20):
\begin{itemize}
    \item Overly conservative conclusions
    \item Mismatched with data support
    \item Vague recommendations
\end{itemize}

\item Content (25/45):
\begin{itemize}
    \item Lacks analysis of abnormal movement causes
    \item Disconnected technical analysis
    \item Superficial multi-dimensional analysis
\end{itemize}

\item Expression (10/15):
\begin{itemize}
    \item Complete but weak hierarchical structure
    \item Lacks professional expression
\end{itemize}

\item Data (15/20):
\begin{itemize}
    \item Basic complete data dimensions
    \item Insufficient data support in some dimensions
\end{itemize}
\item Total: 55
\end{itemize}

\section{A complete example of Stocksis}
Due to space limitations, we can only show part of the content of a sample of Stocksis in Table \ref{tab: tuning_dataset_ex} in the main text. We show the complete content of the same sample in Table \ref{tab: tuning_dataset_full} here.
\begin{table*}[h]
\small
\centering
\begin{tabular}{|p{0.64\textwidth}|p{0.36\textwidth}|}
\hline
\textbf{Prompt with Background Information (input)} & \textbf{Expert-Edited Analysis (label)} \\
\hline
\textbf{System Instruction.}
You are a proficient investment assistant, adept at stock analysis tasks. You demonstrate a robust aptitude for comprehending user-provided \textit{[Question]} and meticulously organizing them into accurate and standardized analysis reports. You utilize the \textit{[Background Information]} provided by the corresponding quantitative tools to respond to users in a systematic and standardized manner.

\vspace{0.2cm}
\textbf{Background Information (get from quantitative tools).}
\begin{itemize}
    \item Volume and Price Analysis: TF Securities is currently trading at 4.48, showing a 0.90\% increase. The stock ranks 896th overall and 8th within the securities industry. The securities sector index stands at 1580.86 with a 0.09\% increase, ranging between a high of 1623.79 and a low of 1553.31, with a total sector trading value of 75.187 billion yuan. While the Shanghai Composite Index is at 3189.89, down 0.36\%, TF Securities demonstrates strong trading activity with a 17.65\% turnover rate (ranking 102nd overall and 1st in the securities industry). On October 16th, the stock recorded an impressive trading volume of 152.947 million shares with a total trading value of 6.970 billion yuan, leading both the overall market and the securities industry in trading volume.
    \item Technical Analysis: A bullish engulfing pattern emerged in TF Securities' chart on September 13th, signaling potential short-term strength. The bullish outlook remains valid as long as the stock maintains support above the pattern's key level of 2.68 yuan.
    \item Capital Flow Analysis: Recent capital flow indicators show mixed signals for TF Securities. While the 5-day cumulative DDX of -14.865 indicates significant selling pressure from major players, the stock maintains medium-level control by market forces with an optimistic outlook. The strong inflow in margin trading and increased institutional holdings quarter-over-quarter suggest positive market sentiment and potential medium-term opportunities.
    \item Financial Analysis: TF Securities released its 2024 semi-annual report on August 31st, which triggered a 23.75\% stock price increase in the subsequent 10 trading days. The report revealed significant decreases in non-recurring net profit, revenue, and net profit attributable to the parent company. However, the Q2 2024 quarterly report showed substantial quarter-over-quarter improvements in these same metrics, indicating a potential turnaround in financial performance.
    \item Fundamental Analysis: The company's fundamental structure heavily relies on fee and commission income, which accounts for 131.22\% of total revenue. When compared to industry peers, TF Securities shows below-average fundamental indicators. As a member of the securities industry according to SWSI classification, the company's growth metrics raise concerns, with negative year-over-year growth rates in revenue, net profit, and earnings per share.
    \item Market Movement Catalysts: Recent market movements have been influenced by several key events. The appointment of Pang Jiemin as the company's Party Secretary has drawn attention, with the Hubei Hongtai Group's Party Secretary and Chairman emphasizing the importance of political awareness and supporting the new leadership. The company subsequently held a cadre meeting to announce this appointment, highlighting commitments to party leadership and annual operational goals. Additionally, the broader market environment has been favorable, with Shanghai and Shenzhen markets' trading volume exceeding 2 trillion yuan, contributing to TF Securities' strong trading performance of 19.4 billion yuan, benefiting from the overall positive momentum in the securities sector.
\end{itemize}

\textbf{Question.}
Please analyze TF Securities.
& 
\textbf{TF Securities has shown unusual market movement due to active market trading, benefiting along with the broader securities sector. In the short term, there are bullish signals warranting attention. However, caution is advised for medium to long-term positions due to weak fundamentals.}

\vspace{0.2cm}
Regarding volume and price, TF Securities' latest stock price is 4.48 yuan, up 0.90\%. Today's turnover rate is 17.65\%, ranking first in the securities industry, with a trading volume of 152.947 million shares and a trading value of 6.970 billion yuan. The stock ranks 8th in its industry for price gains and is outperforming the Shanghai Composite Index, which is down 0.36\%. The volume and price performance show active market participation and high overall engagement.

On the technical front, TF Securities formed a bullish engulfing pattern on September 13th, indicating short-term strength. The bullish outlook remains valid as long as the price maintains above 2.68 yuan. Additionally, institutional analysis tools indicate strong institutional control, suggesting potential opportunities for investors.

From a capital flow perspective, TF Securities' 5-day cumulative DDX of -14.865 indicates significant selling pressure from major players and strong bearish sentiment. Despite medium-level institutional control and determined margin trading inflows suggesting relatively active capital participation, caution is needed regarding short-term capital flows.

Fundamentally, TF Securities' 2024 semi-annual report shows significant decreases in non-recurring net profit, revenue, and net profit attributable to the parent company, indicating poor overall growth. Moreover, the company's heavy reliance on fee and commission income as its main revenue source places it at a disadvantage compared to industry peers.

On the news front, TF Securities recently held a cadre meeting emphasizing strengthened party leadership and commitment to annual operational goals. Furthermore, the Shanghai and Shenzhen markets' trading volume has consistently exceeded 2 trillion yuan, with TF Securities reaching 19.4 billion yuan in trading value, positively impacting the securities sector.

\vspace{0.2cm}
\textbf{In conclusion, TF Securities presents short-term opportunities due to the bullish engulfing pattern in technical analysis. However, investors should maintain caution in medium to long-term positions due to weak fundamentals. Investment decisions should be made with close attention to market trends and company developments.} \\
\hline
\end{tabular}
\caption{A complete example of Stocksis. Same sample as in Table \ref{tab: tuning_dataset_ex}. The average cost of comprehensive analysis written by experts is \$10 per query, with detailed cost breakdowns available in Appendix \ref{sec: cost}.}
\label{tab: tuning_dataset_full}
\end{table*}

In future work, we plan to design detailed prompts that enable LLMs to replace human experts in evaluating analysis reports using the AnalyScore criteria. 

\section{Stocksis Collection and Evaluation Costs}\label{sec: cost}
To further quantify the value of Stocksis and AnalyScore, we disclose the expert-curated Stocksis dataset containing 5,000 entries and the human evaluation costs for assessing model outputs using AnalyScore. Excluding the costs of quantitative tool calls, the expert compilation of comprehensive analyses costs approximately \$75,000 (averaging \$15 per entry). The expert evaluation of 100 outputs from each of the models using AnalyScore criteria cost approximately \$12,000 (averaging \$10 per evaluation across all groups). These figures do not include the additional costs associated with expert development of the AnalyScore evaluation framework.

In future work, we plan to design detailed prompts that enable LLMs to replace human experts in evaluating analysis reports using the AnalyScore criteria. 

\section{Product Release Information}\label{sec: product}
As FinSphere, a powerful stock analysis agent developed by a stock investment advisory company—we currently have a fully functional product demo and have made it freely available to the public in December 2024. Due to double-blind review requirements, we regrettably cannot showcase this promising product to the reviewers at this stage. However, we look forward to including access information for this free public tool in the final version of our paper.

\section{Detailed testing quiries}\label{sec: test_query}
Here we disclose 100 queries used for testing and experts’ scores on FinSphere. For details, please check the Table \ref{tab: test_queery1}, \ref{tab: test_queery2} and \ref{tab: test_queery3}.

\section{Limitation}
FinSphere’s performance depends on the accuracy and availability of real-time financial data, which may impact analysis reliability. The AnalyScore framework still requires human validation, limiting full automation. Additionally, FinSphere may struggle with nuanced financial reasoning and novel market events beyond its training. Future work should focus on improving real-time adaptability, reducing reliance on curated data, and expanding domain coverage for broader financial applications.
\begin{table*}[htbp]
\centering
\begin{tabular}{p{8cm}cccccc}
\hline
Query & Qual & Conc & Cont & Expr & Data & Score \\
\hline
Can you analyze Fueneng Dongfang? & 1 & 10 & 35 & 15 & 15 & 75 \\ 
Please analyze the situation of Oulai New Materials stock. & 1 & 10 & 35 & 15 & 20 & 80 \\ 
Conduct a comprehensive analysis of Kangxi Communications. & 1 & 10 & 35 & 15 & 15 & 75 \\ 
Comprehensive analysis of Chuangyitong. & 0 & 10 & 35 & 15 & 20 & 80 \\ 
Can you provide a detailed interpretation of Lanhai Huanteng stock? & 1 & 10 & 20 & 15 & 20 & 65 \\ 
Please analyze the current status of Guixin Technology stock. & 1 & 10 & 33 & 15 & 15 & 73 \\ 
Is Zhejiang Hengwei worth investing in? Please analyze. & 1 & 10 & 35 & 15 & 20 & 80 \\ 
Conduct a comprehensive analysis of Anshuo Information. & 1 & 10 & 35 & 13 & 15 & 73 \\ 
I am interested in Chenyi Intelligence. Could you analyze it? & 0 & 10 & 10 & 15 & 15 & 50 \\ 
Diagnose Hailun Zhe. & 1 & 10 & 10 & 15 & 20 & 55 \\ 
Can you conduct an in-depth analysis of Fuguang Co., Ltd.? & 1 & 10 & 35 & 15 & 15 & 75 \\ 
Please analyze Haooubo. & 1 & 10 & 5 & 15 & 20 & 50 \\ 
How is Canxin Co., Ltd.? & 1 & 10 & 30 & 15 & 20 & 75 \\ 
Comprehensive analysis of Saiwei Intelligence. & 0 & 10 & 30 & 15 & 20 & 75 \\ 
Please provide a comprehensive evaluation of Cigu Technology. & 0 & 10 & 30 & 15 & 15 & 70 \\ 
Please comment on the overall performance of Sainuo Medical. & 0 & 10 & 10 & 15 & 15 & 50 \\ 
How is Longli Technology? & 1 & 10 & 30 & 15 & 15 & 70 \\ 
Can you provide a comprehensive analysis of Aofu Environmental Protection? & 1 & 10 & 30 & 15 & 15 & 70 \\ 
What is the comprehensive situation of Yubang New Materials? & 1 & 10 & 30 & 15 & 20 & 75 \\ 
Please conduct a comprehensive review of Zhuojin Co., Ltd. & 1 & 10 & 10 & 15 & 15 & 50 \\ 
How is the performance of Huaguang New Materials in all aspects? & 1 & 10 & 5 & 15 & 20 & 50 \\ 
Please conduct a comprehensive analysis of Jiankang. & 1 & 10 & 25 & 15 & 15 & 65 \\ 
Overall analysis of Taifu Pumps. & 1 & 10 & 30 & 15 & 15 & 70 \\ 
Comment on Zhongfu Information. & 1 & 10 & 25 & 15 & 20 & 70 \\ 
Please provide an analysis of Daoshi Technology. & 1 & 10 & 30 & 15 & 20 & 75 \\ 
Comprehensive analysis of Ruisong Technology. & 1 & 10 & 30 & 15 & 20 & 75 \\ 
Comprehensive analysis of Zhongyi Technology. & 0 & 10 & 30 & 15 & 20 & 75 \\ 
Can you provide a comprehensive evaluation and analysis of Aerospace Hongtu? & 1 & 10 & 30 & 15 & 10 & 65 \\ 
Please give specific analysis opinions on Tengjing Technology. & 1 & 10 & 30 & 10 & 20 & 70 \\ 
Comprehensive analysis of Zhenyou Technology. & 1 & 10 & 25 & 15 & 10 & 60 \\ 
How is the overall situation of Huahai Chengke? & 0 & 10 & 35 & 15 & 20 & 80 \\ 
\hline
\end{tabular}
\caption{Testing queries and experts' scores on FinSphere (1/3)}
\label{tab: test_queery1}
\end{table*}

\begin{table*}[htbp]
\centering
\begin{tabular}{p{8cm}cccccc}
\hline
Query & Qual & Conc & Cont & Expr & Data & Score \\
\hline
Can you conduct a comprehensive analysis of Xingqiu Graphite? & 1 & 10 & 35 & 15 & 20 & 80 \\ 
How is the comprehensive analysis of Shanghai Ailu? & 1 & 10 & 30 & 15 & 20 & 75 \\ 
I want to know the details of Taihe Technology. Can you analyze it for me? & 1 & 10 & 30 & 15 & 20 & 75 \\ 
Comprehensive analysis of Dagang Holdings. & 1 & 10 & 25 & 15 & 20 & 70 \\ 
Please analyze Aorui De and provide investment advice. & 1 & 10 & 25 & 15 & 20 & 70 \\ 
Can you analyze Sunshine Real Estate? & 1 & 10 & 25 & 15 & 20 & 70 \\ 
Can you conduct a comprehensive analysis of Hongbaoli? & 1 & 10 & 25 & 15 & 20 & 70 \\ 
Can you analyze Yangzi New Materials for me? & 1 & 10 & 25 & 15 & 20 & 70 \\ 
How has Chunxing Precision performed recently? Can you analyze it? & 1 & 10 & 35 & 15 & 20 & 80 \\ 
Can you provide professional analysis on Tuoshan Heavy Industry? & 1 & 10 & 25 & 15 & 20 & 70 \\ 
I am interested in the analysis of Yayi Technology. Can you share it? & 1 & 10 & 20 & 15 & 20 & 65 \\ 
What are the key points to watch in Bofei Electric? Can you analyze it? & 1 & 10 & 25 & 15 & 20 & 70 \\ 
Can you conduct a detailed analysis of Kangliyuan? & 1 & 10 & 25 & 15 & 20 & 70 \\ 
How comprehensive is the strength of Hope Co., Ltd.? & 1 & 10 & 25 & 15 & 20 & 70 \\ 
Please conduct a comprehensive analysis of Kuntai Co., Ltd. & 1 & 10 & 25 & 15 & 20 & 70 \\ 
Can you provide comprehensive feedback on Taimusi? & 1 & 10 & 25 & 15 & 20 & 70 \\ 
What do you think about Hongming Co., Ltd.? & 1 & 10 & 25 & 15 & 20 & 70 \\ 
Can you look at Wuzhou Medical for me? & 1 & 10 & 25 & 15 & 20 & 70 \\ 
How is Zhejiang Liming recently? & 1 & 10 & 25 & 15 & 20 & 70 \\ 
Please analyze Baolijia, is it good? & 1 & 10 & 25 & 13 & 15 & 63 \\ 
Can you analyze Lvlian Technology? & 1 & 10 & 25 & 13 & 15 & 63 \\ 
Is Shanghai Hejing worth buying? & 1 & 10 & 25 & 13 & 20 & 68 \\ 
What do you think of Qiaoyuan Co., Ltd.? & 1 & 10 & 25 & 15 & 20 & 70 \\ 
How about Zhongji Huanke? & 1 & 10 & 25 & 15 & 20 & 70 \\ 
Can you talk about Kangguan Technology stock? & 1 & 10 & 25 & 15 & 20 & 70 \\ 
Tell me about Guanghe Technology. & 1 & 10 & 25 & 15 & 20 & 70 \\ 
Is Xingchen Technology doing well recently? & 1 & 10 & 25 & 15 & 20 & 70 \\ 
Please interpret Chengdu Huawei stock. & 1 & 10 & 25 & 15 & 20 & 70 \\ 
How about Jinjiang Shipping? & 1 & 10 & 30 & 15 & 20 & 75 \\ 
What about Jinhui Co., Ltd.? & 1 & 10 & 25 & 15 & 20 & 70 \\ 
Can you provide an investment analysis of Dazhu CNC? & 1 & 10 & 25 & 15 & 20 & 70 \\ 
Is Laplace stock good? & 1 & 5 & 20 & 15 & 15 & 55 \\ 
What do you think of Shennong Group stock? & 1 & 10 & 30 & 15 & 20 & 75 \\ 
\hline
\end{tabular}
\caption{Testing queries and experts' scores on FinSphere (2/3)}
\label{tab: test_queery2}
\end{table*}

\begin{table*}[htbp]
\centering
\begin{tabular}{p{8cm}cccccc}
\hline
Query & Qual & Conc & Cont & Expr & Data & Score \\
\hline
Comprehensive analysis of Shichuang Energy. & 1 & 10 & 25 & 15 & 20 & 70 \\ 
Comprehensive analysis of Ningbo Ocean. & 1 & 10 & 25 & 15 & 20 & 70 \\ 
Comprehensive analysis of Longqi Technology. & 1 & 10 & 25 & 15 & 20 & 70 \\ 
Please evaluate Fuerjia Co., Ltd. as a whole. & 1 & 10 & 25 & 15 & 20 & 70 \\ 
Can you provide an overall evaluation of Yongxing Co., Ltd.? & 1 & 10 & 25 & 15 & 20 & 70 \\ 
Comprehensive analysis of Hekeda Co., Ltd. & 1 & 10 & 25 & 15 & 20 & 70 \\ 
Comprehensive analysis of Craftsman Home. & 1 & 10 & 25 & 15 & 20 & 70 \\ 
Comprehensive analysis of International Composite Materials. & 1 & 10 & 25 & 15 & 20 & 70 \\ 
Please diagnose and analyze Suzhou Tianmai. & 0 & 10 & 20 & 15 & 15 & 60 \\ 
Can you diagnose the stock status of Weidian Physiology? & 0 & 10 & 25 & 15 & 20 & 70 \\ 
Please diagnose Weidao Nano. & 1 & 10 & 25 & 15 & 20 & 70 \\ 
Can you diagnose stock 6912? & 0 & 10 & 20 & 15 & 15 & 60 \\ 
How is Hualan Vaccine stock? & 1 & 10 & 25 & 15 & 20 & 70 \\ 
Please diagnose Mousse Co., Ltd.'s stock comprehensively. & 1 & 10 & 25 & 15 & 20 & 70 \\ 
Can you conduct an in-depth analysis of Huabao New Energy stock? & 1 & 10 & 25 & 15 & 20 & 70 \\ 
Please diagnose Haikan Co., Ltd.'s stock. & 1 & 10 & 25 & 15 & 20 & 70 \\ 
Can you provide a professional analysis of Hongsheng Huayuan? & 1 & 10 & 25 & 15 & 20 & 70 \\ 
Please diagnose China Eastern Airlines. & 1 & 10 & 30 & 15 & 20 & 75 \\ 
Can you analyze the stock of Huali Group in detail? & 1 & 10 & 25 & 15 & 20 & 70 \\ 
Please analyze Postal Savings Bank. & 1 & 10 & 25 & 15 & 20 & 70 \\ 
How to analyze the market trend of ICBC? & 1 & 10 & 30 & 15 & 20 & 75 \\ 
How about Guizhou Moutai stock? & 1 & 10 & 30 & 15 & 20 & 75 \\ 
Please analyze Agricultural Bank of China stock. & 1 & 10 & 25 & 15 & 20 & 70 \\ 
Can you analyze China Construction Bank in detail? & 1 & 10 & 30 & 15 & 20 & 75 \\ 
China Petroleum, diagnose it. & 1 & 10 & 30 & 15 & 20 & 75 \\ 
Can you conduct a comprehensive analysis of China Mobile? & 0 & 10 & 25 & 15 & 20 & 70 \\ 
Is China Bank suitable for long-term holding? & 1 & 10 & 30 & 15 & 20 & 75 \\ 
Please analyze China Life Insurance stock. & 1 & 10 & 30 & 15 & 20 & 75 \\ 
Please research Ningde Times stock. & 1 & 10 & 30 & 15 & 20 & 75 \\ 
Please analyze Zhaosheng Micro stock. & 1 & 10 & 25 & 15 & 20 & 70 \\ 
Please analyze Xinda Securities stock. & 1 & 10 & 25 & 15 & 20 & 70 \\ 
Analyze BAIC Blue Valley. & 1 & 10 & 33 & 15 & 20 & 78 \\ 
How to view COSCO Energy stock? & 1 & 10 & 25 & 15 & 20 & 70 \\ 
Analyze Kelun Pharmaceutical stock. & 1 & 10 & 25 & 15 & 20 & 70 \\ 
Can you conduct a comprehensive diagnosis of New Industries stock? & 1 & 10 & 30 & 15 & 20 & 75 \\ 
Comprehensive analysis of Shengyi Technology. & 1 & 10 & 25 & 15 & 20 & 70 \\
\hline
\end{tabular}
\caption{Testing queries and experts' scores on FinSphere (3/3)}
\label{tab: test_queery3}
\end{table*}

\clearpage
\bibliographystyle{named}
\bibliography{ijcai25}

\end{document}